\pdfoutput=1
\documentclass[11pt]{article}

\usepackage{EMNLP2023}

\usepackage{times}
\usepackage{latexsym}

\usepackage[T1]{fontenc}
\usepackage[utf8]{inputenc}

\usepackage{microtype}

\usepackage{inconsolata}
\usepackage{graphicx}
\usepackage{amsmath}
\usepackage{booktabs}
\usepackage{multirow}
\usepackage{subfigure}
\usepackage{appendix}
\usepackage{tablefootnote}

\newcommand{\tabincell}[2]{\begin{tabular}{@{}#1@{}}#2\end{tabular}}

\newcommand{\eat}[1]{}
\newcommand{\eg}{{\em e.g.,~}}      % e.g.
\newcommand{\ie}{{\em i.e.,~}}      % i.e.
\usepackage{color}
\usepackage{framed}
\usepackage{tcolorbox}
\usepackage{wrapfig}

\title{Document-level Claim Extraction and Decontextualisation for Fact-Checking}

\author{
Zhenyun Deng,
Michael Schlichtkrull,
Andreas Vlachos \\
Department of Computer Science and Technology, University of Cambridge \\
\{zd302, mss84, av308\}@cam.ac.uk 
\\
}

\begin{document}
\maketitle
\begin{abstract}
Selecting which claims to check is a time-consuming task for human fact-checkers, especially from documents consisting of multiple sentences and containing multiple claims. However, existing claim extraction approaches focus more on identifying and extracting claims from individual sentences, \eg identifying whether a sentence contains a claim or the exact boundaries of the claim within a sentence. In this paper, we propose a method for \textit{document-level} claim extraction for fact-checking, which aims to extract check-worthy claims from documents and decontextualise them so that they can be understood out of context. Specifically, we first recast claim extraction as extractive summarization in order to identify central sentences from documents, then rewrite them to include necessary context from the originating document through sentence decontextualisation. Evaluation with both automatic metrics and a fact-checking professional shows that our method is able to extract check-worthy claims from documents more accurately than previous work, while also improving evidence retrieval.
\end{abstract}

\section{Introduction}
Human fact-checkers typically select a claim in the beginning of their day to work on for the rest of it. Claim extraction (CE) is an important part of their work, as the overwhelming volume of claims in circulation means the choice of \textit{what} to fact-check greatly affects the fact-checkers' impact~\cite{konstantinovskiy2021toward}.
Automated approaches to this task have been proposed to assist them in selecting check-worthy claims, \ie claims that the public has an interest in knowing the truth~\cite{hassan2017toward,guo2022survey}. 

Existing CE methods mainly focus on detecting whether a sentence contains a claim~\cite{reddy2021newsclaims,nakov2021clef} or the boundaries of the claim within a sentence~\cite{wuhrl2021claim,sundriyal2022empowering}. In real-world scenarios though, claims often need to be extracted from documents consisting of multiple sentences and containing multiple claims, not all of which are relevant to the central idea of the document, and verifying all claims manually or even automatically would be inefficient. 

\begin{figure*}[!ht]
	\begin{center}
		{\scalebox{0.5}              
            {\includegraphics{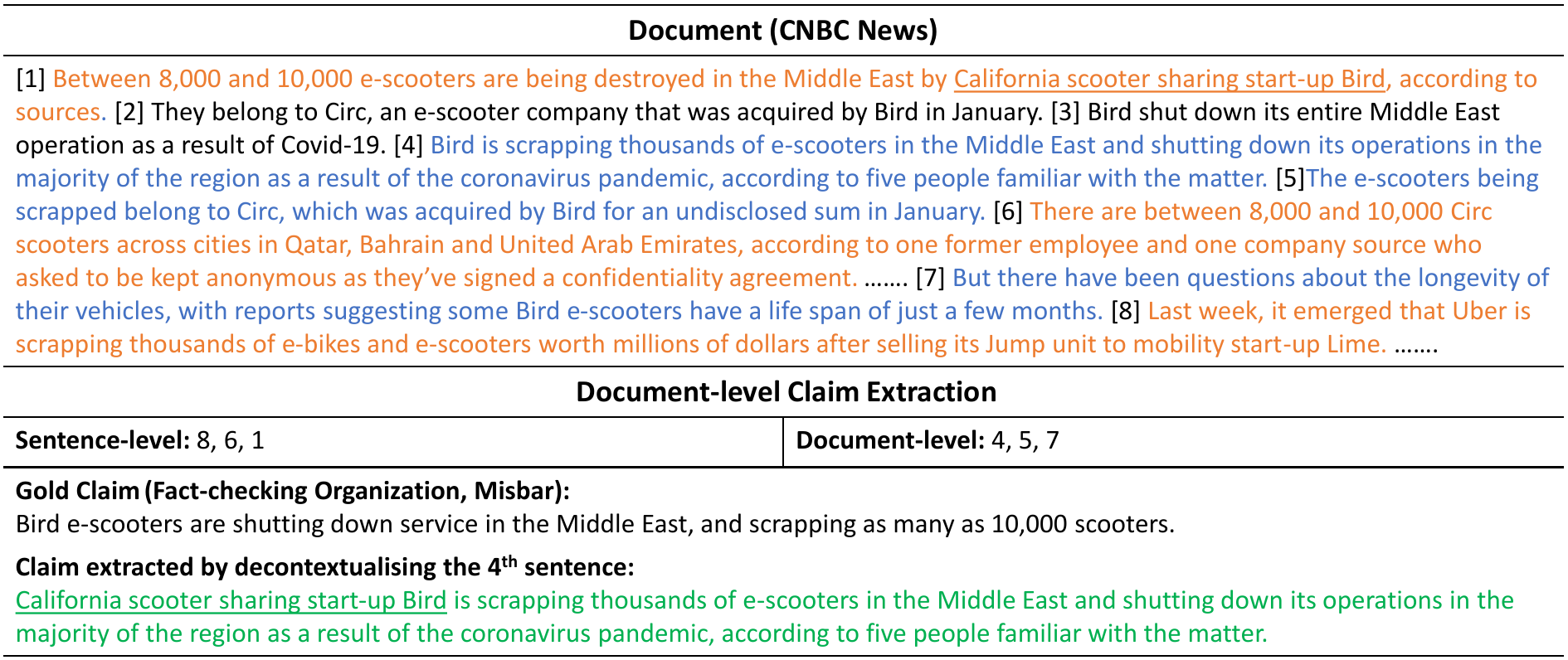}}}
		\caption{ An example of document-level claim extraction. Document\protect\footnotemark[1] is a piece of news from CNBC. Gold Claim\protect\footnotemark[2] is annotated by the fact-checking organization, Misbar. Sentences in orange denote check-worthy claims extracted by sentence-level CE (Claimbuster). Sentences in blue denote salient claims extracted by our document-level CE. The claim in green is a decontextualised claim derived from the 4th sentence obtained by our document-level CE.}
		\label{fig1}
	\end{center}
	\vspace{-5mm}
\end{figure*}

Moving from sentence-level CE to document-level CE is challenging; we illustrate this with the example in Figure~\ref{fig1}. Sentences in orange are claims selected by a popular sentence-level CE method, Claimbuster~\cite{hassan2017claimbuster}, that are worth checking in principle but do not always relate to the central idea of the document, and multiple sentences with similar claims are selected, which would not all need to be fact-checked (\eg sentences 1 and 6).

Claims extracted for fact-checking are expected to be unambiguous~\cite{lippi2015context,wuhrl2021claim}, which means that they cannot be misinterpreted or misunderstood when they are considered outside the context of the document they were extracted from, consequently allowing them to be  fact-checked more easily~\cite{schlichtkrull2023averitec}. Figure~\ref{fig1} shows an example of claim decontextualisation, where the claim ``\textit{Bird is scrapping thousands of e-scooters in the Middle East ...... }'' requires coreference resolution to be understood out of context, \eg ``\textit{Bird}'' refers to ``\textit{California scooter sharing start-up Bird}''. However, existing CE methods primarily focus on extracting sentence-level claims (\ie extracting sentences that contain a claim) from the original document~\cite{reddy2021newsclaims} and ignore their decontextualisation, resulting in claims that are not unambiguously understood and verified.

To address these issues, we propose a novel method for \textit{document-level} claim extraction and decontexualisation for fact-checking, aiming to extract salient check-worthy claims from documents that can be understood outside the context of the document. Specifically, assuming that salient claims are derived from central sentences, $i)$ we recast the document-level CE task into the extractive summarization task to extract central sentences and reduce redundancy; $ii)$ we decontextualise central sentences to be understandable out of context by enriching them with the necessary context; $iii)$ we introduce a QA-based framework to obtain the necessary context by resolving ambiguous information units in the extracted sentence. % into declarative sentences.

\footnotetext[1]{\url{https://www.cnbc.com/2020/06/03/bird-circ-scooters-middle-east.html}}
\footnotetext[2]{\url{https://misbar.com/en/factcheck/2020/06/18/are-bird-e-scooters-leaving-the-middle-east}}
\footnotetext[3]{\url{https://github.com/Tswings/AVeriTeC-DCE}}

To evaluate our method we derive a CE dataset\protect\footnotemark[3] containing decontextualised claims from AVeriTeC \cite{schlichtkrull2023averitec}, a recently proposed benchmark for real-world claim extraction and verification.
% To evaluate our method we derive a CE dataset\protect\footnotemark[3] containing decontextualised claims from AVeriTeC \cite{schlichtkrull2023averitec}, a recently proposed dataset containing claims from real-world fact-checking articles. 
Our method achieves a Precision@1 score of 47.8 on identifying central sentences, a 10$\%$ improvement over Claimbuster. This was verified further by a fact-checking professional, as the sentences returned by our method were deemed central to the document more often, and check-worthy more often than those extracted by Claimbuster. Additionally, our method achieved a character-level F score  (chrF)~\cite{popovic2015chrf} of 26.4 against gold decontextualised claims, 
%by the fact-checkers, 
outperforming all baselines. When evaluated for evidence retrieval potential, the decontextualised claims obtained by enriching original sentences with the necessary context, are better than the original claim sentences, with an average 1.08 improvement in precision.

\begin{figure*}
	\begin{center}
		{\scalebox{0.50}  
            {\includegraphics{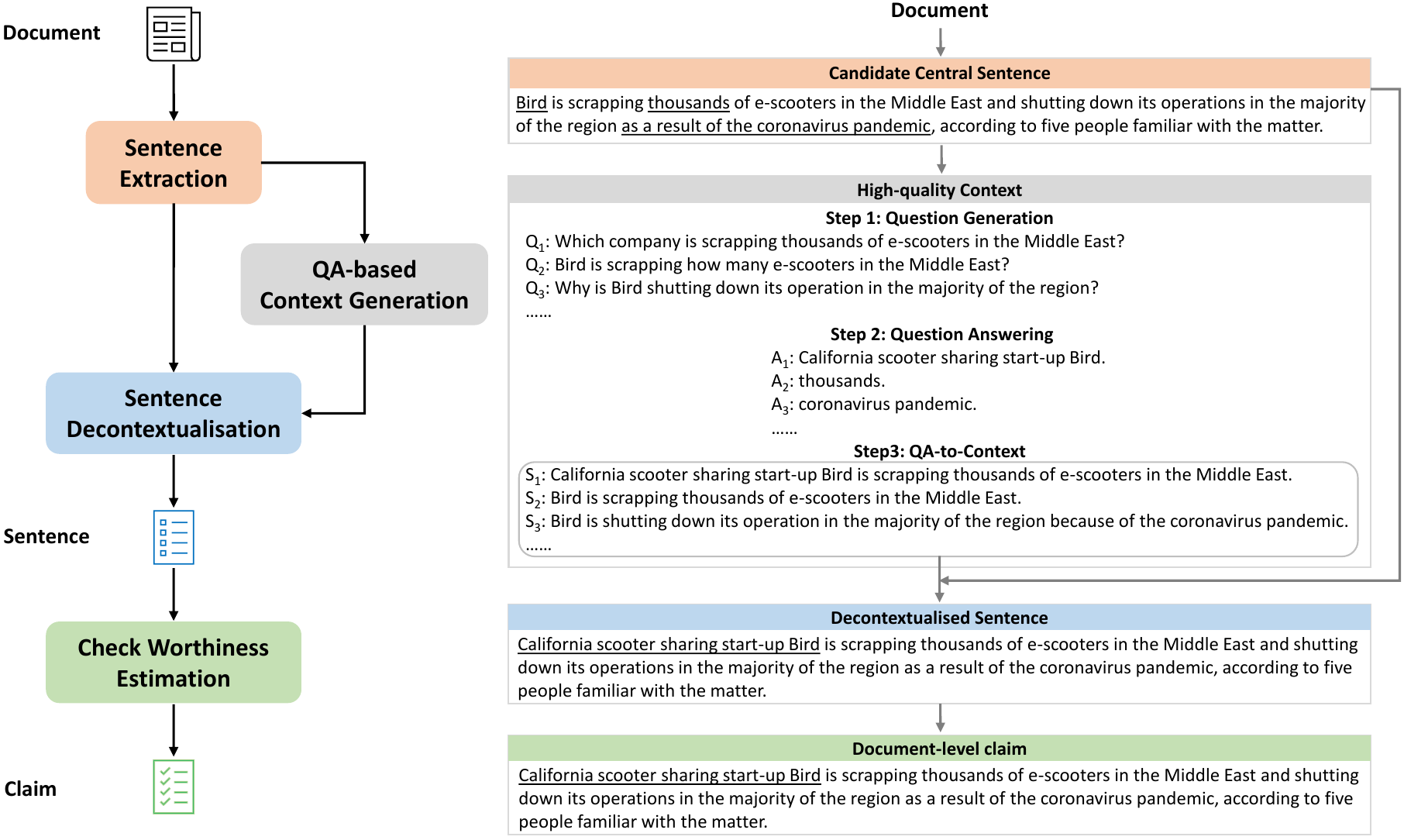}}}
		\caption{
		An overview of our document-level claim extraction framework. Given an input document, we first use extractive summarization to rank all sentences and select summary sentences as central sentences. Then, we describe a QA-based framework to generate a specific high-quality context for important information units in the sentence. Next, we use a seq2seq generation model to decontextualise sentences by enriching them with their corresponding context. Finally, a claim check-worthiness classifier is used to select salient check-worthy claim sentences based on the score that reflects the degree to which sentences belong to the check-worthy claim.
        }
		\label{framework}
	\end{center}
	\vspace{-1mm}
\end{figure*}

\section{Related Work}
\paragraph{Claim Extraction} Claim extraction is typically framed either as a classification task or claim boundary identification task. The former framing focuses on detecting whether a given sentence contains a check-worthy claim. Claimbuster~\cite{hassan2017claimbuster}, the most popular method in this paradigm, computes the score of how important a sentence is to be fact-checked. 
% \citet{nakov2021overview} formulate the check-worthy claim detection task as the sentence ranking task, and then select the sentences at the top of the list as claim sentences.
More similar to our work are studies that formulate the task of check-worthy claim detection as a sentence ranking task. For example, \citet{zhou2021fight} present a sentence-level classifier by combining a fine-tuned hate-speech model with one dropout layer and one classification layer to rank sentences. However, these methods were not able to handle the challenges of document-level claim extraction, \eg avoid redundant claim sentences. 

The framing of claim extraction as boundary identification focuses on detecting the exact claim boundary within the sentence. \citet{nakov2021automated} propose a BERT-based model to perform  claim detection \cite{levy2014context} by identifying the boundaries of the claim within the sentence. \citet{sundriyal2022empowering} tackle claim span identification as a token classification task for identifying argument units of claims in the given text. Unlike the above methods, where the claims are extracted from given sentences, our work aims to extract salient check-worthy claims from documents, thus addressing the limitations of sentence-level methods in extracting salient claim sentences and avoiding redundancy.

\paragraph{Decontextualisation}
%decontextualisation is the process of enriching a sentence with its necessary context making it to be understandable out of context while retaining its original meaning. 
\citet{choi2021decontextualization} propose two different methods for decontextualisation, based on either a coreference resolution model or a seq2seq generation model. Both methods use the sentences in the paragraph containing the target sentence as context to rewrite it. \citet{newman2023controllable} utilize an LLM to generate QA pairs for each sentence by designing specific prompts, and then use an LLM with these QA pairs to rewrite each sentence. \citet{sundriyal2023chaos} propose to combine chain-of-thought and in-context learning for claim normalization. Unlike the above methods, we generate declarative sentences for potentially ambiguous information units in the target sentence based on the whole document, and combine them into context to rewrite the target sentence. %, thereby improving the sensitivity to ambiguity when decontextualising.

\section{Method}
As illustrated in Figure~\ref{framework}, our proposed document-level claim extraction framework consists of four components: $i)$ Sentence extraction ($\S 3.1$); extracts the sentences related to the central idea of the document as candidate claim sentences; $ii)$ Context generation ($\S 3.2$), extracts context from the document for each candidate sentence; $iii)$ Sentence decontextualisation ($\S 3.3$), rewrites each sentence with its corresponding context to be understandable out of context; $iv)$ Check-worthiness estimation ($\S 3.4$), selects the final check-worthy claims from candidate decontextualised sentences.

\subsection{Sentence Extraction} \label{section_3_1}
The claims selected by human fact-checkers are typically related to the central idea of the document considered. Thus we propose to model sentence extraction as extractive summarization. For this purpose, we concatenate all the sentences in the document into an input sequence, 
which is then fed to BertSum~\cite{liu2019text}, a document-level extractive summarization method trained on the CNN/DailyMail dataset. %% that can express the semantics of a document, to extract central sentences. 
Specifically, given a document consisting of $n$ sentences $\mathcal{D}=\{{s}_1, {s}_2, ..., {s}_n\}$, we first formulate the input sequence $C$ as ``${\rm [CLS]}\ {s}_1\ {\rm [SEP]} \ {\rm [CLS]}\ {s}_2$ ${\rm [SEP]}\ ...\ {\rm [CLS]} \ {s}_n {\rm [SEP]}$'', where $\rm [CLS]$ and $\rm [SEP]$ denote the start and end token for each sentence, respectively, and then feed them into a pre-trained encoder BERT to obtain the sentence representation $\mathbf s$. Finally, a linear layer on sentence representations $\mathbf{S}  = \{\mathbf{s}_1, ..., \mathbf{s}_i, ...,  \mathbf{s}_n\}$ is used to score sentences.
\begin{eqnarray}
\begin{array}{l}
\begin{aligned}
\mathbf{S} & = {\rm BERT}(C)    \\
score_i & = \sigma(W\mathbf{s}_i+b_0)
% Score = \{score_i\}^{n}_{i=1} & = \{\sigma(W\mathbf{s}_i+b_0)\}^{n}_{i=1}
\end{aligned}
\end{array}
\label{eq3_1_1}
\end{eqnarray}
where $\sigma$ is a sigmoid function, $\mathbf{s}_i$ denotes the representation of the $i$-th $\rm [CLS]$ token, \ie the representation of the $i$-th sentence, and $score_i$ denotes the score of the $i$-th sentence. All sentences are constructed into an ordered set $S = \{s'_1, ..., s'_i, ..., s'_n\}$ according to their scores. Since all sentences are ranked by sentence-level scoring, some top-scoring sentences may have the same or similar meaning. To avoid redundancy, we add an entailment model DocNLI~\cite{yin2021docnli}, a more generalizable model trained on five datasets from different benchmarks, on top of the output of BertSum to remove redundant sentences by calculating the entailment scores between sentences, \eg we first remove the sentences that have an entailment relationship with the top-1 sentence in $S$, and then repeat this process for the remaining top-2/3/... sentence until we extract $k$ central sentences. Following previous work~\cite{liu2019text}, we only select the top-$k$ sentences with the highest scores in Equation~\ref{eq3_1_1} as candidate central sentences.
% \eg if candidate sentences have an entailment relationship with the top-1 sentence, we remove them; we then repeat this process for the top-2/3/... sentence until we extract $k$ central sentences. Following previous work~\cite{liu2019text}, we only select the top-$k$ sentences with the highest scores in Equation~\ref{eq3_1_1} as candidate central sentences.
\begin{eqnarray}
\begin{array}{l}
\begin{aligned}
S' & = {\rm DocNLI}(S)
\end{aligned}
\end{array}
\label{eq3_1_2}
\end{eqnarray}
where $S'=\{s'_1, s'_2, ..., s'_k\}$ is a set of central sentences that do not contain the same meaning.

\subsection{Context Generation}
After sentence extraction, the next step is to clarify the (possibly) ambiguous sentences in $S'$ by rewriting them with their necessary context. Unlike \citet{choi2021decontextualization} where the context consists of a sequence of sentences in the paragraph containing the ambiguous sentence, we need to consider the whole document, \ie sentences from \textit{different} paragraphs. 
%To improve the sensitivity to ambiguity when decontextualising, 
We propose a QA-based context generation framework to produce a specific context for each ambiguous sentence, which contains three components: $i$) Question Generation: extracts potentially ambiguous information units from the sentence and generates questions with them as answers; $ii$) Question Answering: finds more information about ambiguous information units by answering generated questions with the whole document; $iii$) QA-to-Context Generation: converts question-answer pairs into declarative sentences and combines them into context specific to the sentence. In the following subsections, we describe each component in detail.

\paragraph{Question Generation.} To identify ambiguous information units in candidate central sentences, we first use Spacy\protect\footnotemark[4] to extract named entities, pronouns, nouns, noun phrases, verbs and verb phrases in the sentence $i$ as the potentially ambiguous information units ${U}_i=$ $\{u^1_i, u^2_i, ..., u^j_i, ..., u^m_i\}, i \in [1,2,...,k]$, where $u^j_i$ denotes the $j$-th information unit of the $i$-th candidate sentence $s'_i$.

\footnotetext[4]{\url{https://spacy.io}}

Once the set of information units for a sentence ${U}_i$ is identified, we then generate a question for each of them. Specifically, we concatenate $u^j_i$ and $s'_i$ in which $u^j_i$ is located as the input sequence and feed it into $\rm QG$~\cite{murakhovs2021mixqg}, a question generator model trained on nine question generation datasets with different types of answers, to produce the question $q^j_i$ with $u^j_i$ as the answer. 
\begin{eqnarray}
\begin{array}{l}
{Q}_i = \{q^j_i\}^{m}_{j=1} = \{{\rm QG}(s'_i, u^j_i)\}^{m}_{j=1}
\end{array}
\label{eq3_2_1}
\end{eqnarray}
where ${Q}_i$ denotes the set of questions corresponding to $U_i$ in the sentence $s'_i$.

\begin{table*}
	\centering
	\begin{tabular} {l |c |c |c |c |c} \toprule
            \tabincell{c}{Data}  &\#$\rm sample$ &$\rm med.sent$ &$\rm avg.sent$ &$\rm len.claim$ &$\rm len.document$ \\  \midrule 
            Train &830  &9 &1.09 &17 &274  \\
 		Dev   &149  &6 &1.01 &16 &120  \\
		  Test  &252  &5 &1.05 &16 &63 \\ \midrule 
		All   &1231 &7 &1.07 &17 &17 \\
  \bottomrule
	\end{tabular}
	\caption{ Descriptive statistics for AVeriTeC-DCE. \#$\rm sample$ refers to the number of samples in AVeriTeC available for claim extraction, \ie the total number of accessible $source \ url$, $\rm med.sent$ refers to the median number of sentences in documents. $\rm avg.sent$ refers to the average number of sentences in claims, $\rm len.claim$ refers to the median length of claims in words, $\rm len.document$ refers to the median length of documents in words.}
	\label{tab1}
        \vspace{-3mm}
\end{table*}

\paragraph{Question Answering.} After question generation, our next step is to clarify ambiguous information units by answering corresponding questions with the document $\mathcal{D}$. Specifically, following \citet{schlichtkrull2023averitec}, %to maintain consistency with AVeriTeC,
we first use BM25~\cite{robertson2009probabilistic} to retrieve evidence $E$ related to the question $q^j_i$ from $\mathcal{D}$, and then answer $q^j_i$ with $E$ using an existing QA model~\cite{khashabi2022unifiedqa} trained on twenty datasets that can answer different types of questions.
% using an existing QA model~\cite{khashabi2022unifiedqa}:
\begin{eqnarray}
\begin{array}{c}
\begin{aligned}
E  &= {\rm BM25}(\mathcal{D},q^j_i) \\
a^j_i & = {\rm QA}(E,q^j_i)
\end{aligned}
\end{array}
\label{eq3_2_2}
\end{eqnarray}
where $a^j_i$ denotes a more complete information unit corresponding to ${u}^i_j$, \eg a complete coreference. We denote all question-answer pairs of the $i$-th sentence as $P_i=\{(q^1_i, a^1_i), (q^2_i, a^2_i), ..., (q^m_i, a^m_i)\}$.

\paragraph{QA-to-Context Generation.} After question answering, we utilize a seq2seq generation model to convert QA pairs $P_i$ into the corresponding context $C'_i$. Specifically, we first concatenate the question $q^j_i$ and the answer $a^j_i$ as the input sequence, and then output a sentence using the BART model \cite{lewis2019bart} finetuned on QA2D \cite{demszky2018transforming}. QA2D is a dataset with over 500k NLI examples that contains various inference phenomena rarely seen in previous NLI datasets. More formally,
\begin{eqnarray}
\begin{array}{l}
\tilde{s}^j_i = {\rm BART}(q^j_i, a^j_i)
\end{array}
\label{eq3_2_3}
\end{eqnarray}
where $\tilde{s}^j_i$ is a declarative sentence corresponding to the information unit ${u}^j_i$. Finally, all generated sentences are combined into high-quality context $C'_i=\{\tilde{s}^1_i, \tilde{s}^2_i, ..., \tilde{s}^m_i\}$ corresponding to the information units ${U}_i$ in sentence $s'_i$, which is then used in the next decontextualisation step to enrich the ambiguous sentences.

\subsection{Sentence Decontextualisation}
Sentence decontextualisation aims to rewrite sentences to be understandable out of context, while retaining their original meaning. To do this, we use a seq2seq generation model T5 \cite{raffel2020exploring} to enrich the target sentence with %its corresponding high-quality
the 
context generated in the previous step for it. Specifically, 
we first formulate the input sequence as ``$ {\rm [CLS]}\ \tilde{s}^1_i \ {\rm [SEP]} \ \tilde{s}^2_i \ ...... \ {\rm [SEP]}\ \tilde{s}^m_i \ {\rm [SEP]}\ s'_i$'', where $s'_i$ denotes the potential ambiguous sentence and $\rm [SEP]$ is a separator token between the context sentences generated. We then feed the input sequence to $\rm D$ \cite{choi2021decontextualization}, a decontextualisation model was trained on the dataset annotated by native speakers of English in the U.S.\ that handles various linguistic phenomena, to rewrite the sentence. Similarly, we set the output sequence to be ${\rm [CAT]\ [SEP]}\ y$.
\begin{eqnarray}
\begin{array}{l}
{y_i}=\left\{
\begin{aligned}
{\rm D}({s'_i, C'_i}), &~if~{{\rm CAT}=feasible} \\
~~~~~{s'_i}~~~~~, &~if~{{\rm CAT}=infeasible} \\
~~~~~{s'_i}~~~~~, &~if~{{\rm CAT}=unnecessary}
\end{aligned}
\right.
\end{array}
\label{eq3_3}
\end{eqnarray}
where ${\rm CAT}=feasible$ or $infeasible$ denotes that $s'_i$ can or cannot be decontextualised, ${\rm CAT} = unnecessary$ denotes that $s'_i$ can be understood without being rewritten, $y_i$ denotes the $i$-th decontextualised sentence.

\subsection{Check-Worthiness Estimation}
Unlike existing CE methods that determine whether a sentence is worth checking without considering the context, we estimate the check-worthiness of a sentence after decontextualisation because some sentences may be transformed from not check-worthy into check-worthy ones in this process. Specifically, we use a DeBERTa model trained on the ClaimBuster dataset \cite{arslan2020benchmark} to classify sentences into three categories: Check-worthy Factual Sentence (CFS), Unimportant Factual Sentence (UFS) and Non-Factual Sentence (NFS). Formally, 
\begin{eqnarray}
\begin{array}{c}
score(y_i) = {\rm DeBERTa}(class={\rm CFS}\ | \ y_i)    \\
{\rm claim} = {\rm argmax}\{score(y_i)\}^{k}_{i=1}
\end{array}
\label{eq3_4}
\end{eqnarray}
where $score(y_i)$ reflects the degree to which the decontextualised sentence $y_i$ belongs to CFS, and $\rm claim$ denotes that the final salient check-worthy claim that can be understood out of context.

\section{Dataset}
We convert AVeriTeC, a recently proposed dataset for real-world claim extraction and verification \cite{schlichtkrull2023averitec}, into AVeriTeC-DCE, a dataset  for the document-level CE task.
AVeriTeC is collected from 50 different fact-checking organizations and contains 4568 real-world claims. Each claim is associated with attributes such as its type, source and date. In this work, we mainly focus on the following attributes relevant to claim extraction: $i$) $claim$, the claim as extracted by the fact-checkers and decontextualised by annotators, and $ii$) $source \ url$: the URL linking to the original web article of the $claim$. The task of this work is to extract the salient check-worthy claims from the $source \ url$. We also consider whether claims need to be decontextualised when extracting them from documents, as this will directly affect the subsequent evidence retrieval and claim verification. 

To extract claim-document pairs from AVeriTeC that can be used for document-level CE, we perform the following filtering steps: $1)$ Since we focus on the extraction of textual claims, we do not include $source \ urls$ containing images, video or audio; $2)$ To extract the sentences containing the claims from $source \ urls$, we build a web scraper to extract text data in the $source \ url$ as the document. We found that the attribute $source \ url$ is not always available in samples, thus we only select those samples where the web scraper can return text data from $source \ urls$. We obtain a dataset AVeriTeC-DCE, containing 1231 available samples, for document-level CE. We do not divide the dataset into train, dev and test sets, as all models we rely on are pre-trained models (\eg BertSum) and approaches that do not require training (\eg BM25). Statistics for AVeriTeC-DCE are described in Table~\ref{tab1}.

\section{Experiments}
Our approach consists of four components: sentence extraction, context generation, sentence decontextualisation and check-worthiness estimation. As such, 
we conduct separate experiments to evaluate them, as well as an overall evaluation for document-level CE.

\subsection{Sentence Extraction} \label{sec_5_1}
We compare our sentence extraction method, the combination of BertSum and DocNLI stated in Section~\ref{section_3_1}, against other baselines through automatic evaluation and human evaluation.

\paragraph{Baselines}
1) Lead sentence: the lead (first) sentence of most documents is considered to be the most salient, especially in news articles \cite{narayan2018don}; 2) Claimbuster \cite{hassan2017claimbuster}: we use this well-established method to compute the check-worthiness score of each sentence and we rank sentences based on their scores; 3) LSA \cite{gong2001generic}: a common method of identifying central sentences of the document using the latent semantic analysis technique; 4) TextRank \cite{mihalcea2004textrank}: a graph-based ranking method for identifying important sentences in the document; 5) BertSum \cite{liu2019text}: a BERT-based document-level extractive summarization method for ranking sentences.

\paragraph{Automatic Evaluation}
Since the central sentences of documents are not given in AVeriTeC, we cannot evaluate the extracted central sentences by exact matching. Thus, we instead rely on the sentence that has the highest chrF \cite{popovic2015chrf} with the $claim$, as the $claim$ is the central claim annotated by human fact-checkers. We use Precision@k as the evaluation metric, which denotes the probability that the first k sentences in the extracted sentences contain the central sentence. Table~\ref{tab2} shows the results of different sentence extraction methods. We can see that our method outperforms all baselines in identifying the central sentence, achieving a P@1/P@3/P@5/P@10 score of 47.8/63.1/68.6/73.8, which indicates that the combination of the extractive summarization (BertSum) and entailment model (DocNLI) can better capture the central sentences and avoid redundant ones. We found that the common extractive summarization methods (\eg Lead Sentence, TextRank and BertSum) are better than Claimbuster on P@1, confirming what we had stated in the introduction, that sentence-level CE methods have limitations when they are applied at the document-level CE. Moreover, we observe that the lead sentence achieves a P@1 score of 42.3, indicating that there is a correlation between the sentences selected for fact-checking and the lead sentence that often served as the summary. We list the source URLs of the samples for claim extraction in Appendix~\ref{appendix_A1}.

\begin{table}
    % \small
	\centering
	\begin{tabular} {l |c |c |c |c} \toprule
            \multirow{1}{*}{Method} 
            &P@1 &P@3 &P@5 &P@10 \\  \midrule 
            Claimbuster   &37.8  &59.1  &65.7 &71.4 \\  \midrule
            Lead Sentence  &42.3  &-     &-    &-   \\
		  % KL-Sum        &36.5  &54.9  &61.1 &67.5  \\
		LSA           &38.4  &55.3  &62.1 &70.4 \\
		  TextRank      &42.7  &60.6  &65.1 &71.2  \\ 
		BertSum       &43.4  &61.6  &67.5 &72.3  \\    \midrule 
            Ours         &\bf47.8  &\bf63.1  &\bf68.6 &\bf73.8 \\ 
  \bottomrule
	\end{tabular}
	\caption{ Results with different sentence extraction methods. P@k denotes the probability that the first k sentences in the ranked sentences contain the central sentence.}
	\label{tab2}
        % \vspace{-3mm}
\end{table}

\begin{table}
    % \small
	\centering
	\begin{tabular} {l| c |c} \toprule
		\tabincell{c}{}  & Claimbuster  &Ours \\  \midrule       
		  IsCheckWorthy   &0.36  &0.44  \\
		IsCentralClaim  &0.24  &0.68  \\
  \bottomrule
	\end{tabular}
	\caption{ Human Evaluation of sentence extraction on two different dimensions.}
	\label{tab3}
        \vspace{-3mm}
\end{table}

\begin{figure*}
	\begin{center}
		\subfigure{\scalebox{0.5} {\includegraphics{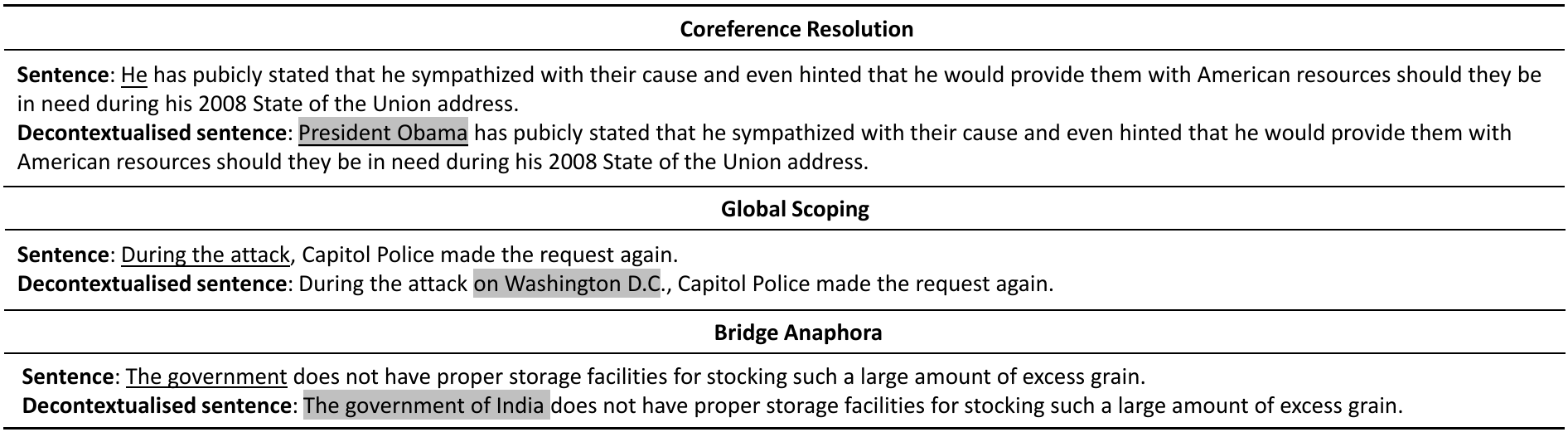}}}
		\vspace{-4mm}
		\caption{ Case studies of sentence decontextualisation solving linguistic problems, such as coreference resolution, global scoping and bridge anaphora.}
		\label{fig3}
	\end{center}
        \vspace{-3mm}
\end{figure*}

\paragraph{Human Evaluation} To further compare sentences extracted by our method and Claimbuster, we asked a fact-checking professional to evaluate the quality of extracted sentences on the following two dimensions: 1) $\bf IsCheckWorthy$: is the sentence worth checking? 2) $\bf IsCentralClaim$: is the sentence related to the central idea of the article? We randomly select 50 samples, each containing at least 5 sentences. For simplicity, we only select the top-1 sentence returned by each method for comparison. As shown in Table~\ref{tab3}, we observed that 68$\%$ of the central sentences extracted by our method are related to the central idea of the document compared to 24$\%$ of Claimbuster, which further supports our conclusion obtained by automatic evaluation, \ie the sentences extracted by our method were more often central to the document, and more often check-worthy than those that extracted by Claimbuster. This indicates that when identifying salient check-worthy claims from documents, it is not enough to consider whether a sentence is worth checking at the sentence level, but also whether the sentence is related to the central idea of the document. Thus, we believe that the claims related to the central idea of the document are the ones that the public is more interested in knowing the truth.

\subsection{Decontextualisation}
To evaluate the effectiveness of decontextualisation on evidence retrieval, for a fair comparison, we select the sentence that has the highest chrF with the $claim$ as the best sentence as we considered in Section~\ref{sec_5_1}, and conduct a comparison between it and its corresponding decontextualised sentence. 
The evidence set used for evaluation is retrieved from the Internet using the Google Search API given a claim, each containing gold evidence and additional distractors \cite{schlichtkrull2023averitec}.
We use Precision@k as the evaluation metric.

\paragraph{Baselines}
1) Coreference model: decontextualisation by replacing unresolved coreferences in the target sentence, \eg \cite{joshi2020spanbert}; 2) Seq2seq model: decontextualisation by rewriting the target sentence with necessary context \cite{choi2021decontextualization}.

\paragraph{Retrieval-based Evaluation}  
For a given claim different decontextualisations could be considered correct, thus comparing against the single reference in AVeriTeC would be suboptimal. Thus we prefer to conduct the retrieval-based evaluation, assuming that better decontextualisation improves evidence retrieval, as it should provide useful context for fact-checking. Following previous work \cite{choi2021decontextualization}, we compare our QA-based decontextualisation method against other baselines through retrieval-based evaluation. We use BM25 as the retriever to find evidence with different sentences as the query. Table~\ref{tab4} shows the results for evidence retrieval with different decontextualised sentences on the dev set of AVeriTeC-DCE (AVeriTeC only publicly released the train and dev sets). We found that our method outperforms all baselines, and improves the P@3/P@5/P@10 score over the original sentence by 1.42/0.82/0.99, achieving an average 1.08 improvement in precision. After further analysis, we found that only 21/149 sentences are decontextualised by our method, and 17/21 of these sentences obtain better evidence retrieval, with an average improvement of 1.21 in precision over the original sentences, proving that decontextualisation enables evidence retrieval more effectively.

\begin{table}[!h]
% \small
\centering
\begin{tabular} {l| c |c | c} \toprule
    \tabincell{c}{Method}  &P@3 &P@5 &P@10 \\  \midrule   
        Sentence              &35.45  &44.72 &61.31 \\ \midrule  
        Coreference           &36.02  &44.98 &61.79 \\ 
        Seq2seq(Context)    &36.17  &45.04 &61.99 \\  
        Ours$_{\rm seq2seq(Context^\star)}$    &\bf36.87 &\bf45.54 &\bf62.30 \\ 
        % Ours$_{\rm seq2seq}$ (Context$^\star$)       &\bf36.87 &\bf45.54 &\bf62.30 \\
\bottomrule
\end{tabular}
\caption{Results for evidence retrieval with different decontextualised sentences. Context consists of a sequence of sentences in the paragraph containing the target sentence. Context$^\star$ consists of declarative sentences generated by our context generation module.}
\label{tab4}
\vspace{-3mm}
\end{table}

\paragraph{Case Study}
Figure~\ref{fig3} illustrates three case studies of sentence decontextualisation. The first case is an example that requires coreference resolution. To make the sentence understandable out of context, these words (\eg ``\textit{He}'', ``\textit{their}'') need to be rewritten with the context. After decontextualisation, we can see that ``\textit{He}'' is rewritten to ``\textit{President Obama}'', which helps us understand the sentence better without context. As for ``\textit{their}'', we cannot decontextualise it because there is no information about this word in the document. This supports the claim that providing a high-quality context is necessary for better decontextualisation. The second case is an example that requires global scoping, which requires adding a phrase (\eg prepositional phrase) to the entire sentence to make it better understood. In this case, we add ``\textit{on Washington D.C.}'' as a modifier to ``\textit{During the attack}'' to help us understand where the attack took place.  The third case is an example that requires a bridge anaphora, where the phrase noun ``\textit{The government}'' becomes clear by adding a modifier ``\textit{India}''. In summary, decontextualising the claim is helpful for humans to better understand the claim without context.

\subsection{Document-level Claim Extraction}
In this section, we put four components together to conduct an overall evaluation for document-level CE, \eg extracting the claim in green from the document in Figure~\ref{fig1}. We first select top-3 sentences returned by different sentence extraction methods as candidate central sentences, and then feed them into our decontextualisation model to obtain decontextualised claim sentences, finally use a claim check-worthiness classifier to select the final claim. We evaluate performance by calculating the similarity between our final decontextualised claim sentence and the $claim$ decontextualised by fact-checkers. We use the chrF as the evaluation metric. The chrF computes the similarity between texts using the character n-gram F-score. Other metrics are reported in Appendix~\ref{appendix_A2}.

\begin{table}[!h]
    % \small
	\centering
	\begin{tabular} {l|c |c |c |c } \toprule
		\tabincell{c}{Data}  &\#$\rm claim$ &\#$\rm fea.$ &\#\rm $\rm 
 infea.$ &\#$\rm unnec.$ \\  \midrule 
            All       &1231  &122 &122 &987 \\
  \bottomrule
	\end{tabular}
	\caption{Statistic of decontextualisation. \#$\rm fea./ infea.$ denotes the number of sentences that can/cannot be decontextualised, \#$\rm unnec.$ denotes the number of sentences that can be understood without context.}
	\label{tab5}
 \vspace{-3mm}
\end{table}

\begin{table}[!h]
    % \small
	\centering
	\begin{tabular} {l |c |c} \toprule
            \multirow{2}{*}{Method} &\multicolumn{2}{c}{chrF} \\
            \cmidrule(lr){2-3}  &Sentence$^\star$  &Dec. Sentence$^\star$ \\ \midrule 
            Claimbuster   &24.3   &24.5 \\  \midrule
            Lead Sentence  &23.8  &-   \\
		LSA           &24.1  &24.3  \\
		  TextRank      &24.5  &25.4  \\ 
		BertSum       &25.6   &25.9  \\    \midrule 
            Ours        &\bf25.9   &\bf26.4\\ 
  \bottomrule
	\end{tabular}
	\caption{Results of Document-level CE. Sentence$^\star$ denotes the best sentence returned by different sentence extraction methods. Dec. Sentence$^\star$ denotes the decontextualised Sentence$^\star$.}
	\label{tab6}
        \vspace{-3mm}
\end{table}

Statistics for decontextualisation are described in Table~\ref{tab5}, we observe that 122 out of 1231 (10$\%$) sentences can be decontextualised (feasible); 122 out of 1231 (10$\%$) sentences cannot be decontextualised (infeasible); and 987 out of 1231 (80$\%$) sentences can be understood without being decontextualised (unnecessary), including the lead sentence. Since the lead sentence of the document is often considered to be the most salient, we do not decontextualise the lead sentence. In Table~\ref{tab6}, we show the results of original sentences and decontextualised sentences for claim extraction, and our method achieves a chrF of 26.4 on gold claims decontextualised by the fact-checkers, outperforming all baselines.
We observe that the performance of claim extraction and sentence extraction is positively correlated, \ie the closer the extracted sentence is to the central sentence, the more similar the extracted claim is to the $claim$, which supports our assumption that salient claims are derived from central sentences. For this reason, the performance of our document-level CE is limited by the performance of sentence extraction, \ie if our sentence extraction method cannot find the gold central sentence, decontextualisation may not improve the performance of CE, and may even lead to a decrease in the performance of CE due to noise caused by decontextualisation. 

Moreover, empirically, we found that central sentences led to improved overall performance. This might be a consequence of the dataset – social media sites such as Twitter or Facebook are common sources of claims in AVeriTeC (along with more traditional news), and a Twitter or Facebook thread often contains only a few key points. AVeriTeC was collected by reverse engineering claims which fact-checkers from around the world chose to work on. As such, the distribution of source articles represents what journalists found to be check-worthy and chose to work on. Our work as such reflects the contexts wherein real-world misinformation appears (but indeed, may have a bias towards what works well in those contexts).

To further verify the effectiveness of our method, we conduct a comparison on the document-level CE dataset (CLEF-2021, subtask 1B \cite{shaar2021overview}) using our method and Claimbuster. Table \ref{tab7} shows the results of two different methods for identifying check-worthy claims on the dev set of subtask 1B. We observe that our method outperforms Claimbuster on P@1/3/5/10, indicating that our document-level CE method can better identify check-worthy claims than Claimbuster (sentence-level CE). This supports our conclusion that the document-level check-worthy claims extracted by our method are the claims that the public is more interested in knowing the truth.

\begin{table}[!h]
% \small
\centering
\begin{tabular} {l| c |c | c | c} \toprule
    \tabincell{c}{Method}  &P@1 &P@3 &P@5 &P@10   \\  \midrule   
        Claimbuster &0.111  &0.074 &0.156 &0.089  \\ 
        Ours        &0.222  &0.185 &0.200 &0.144  \\
\bottomrule
\end{tabular}
\caption{Results of different methods for identifying check-worthy claims on the dev set of subtask 1B.}
\label{tab7}
 \vspace{-3mm}
\end{table}

\section{Conclusions and Future work}
This paper presented a \textit{document-level} claim extraction framework for fact-checking, aiming to extract salient check-worthy claims from documents that can be understood out of context. To extract salient claims from documents, we recast the claim extraction task as the extractive summarization task to select candidate claim sentences. To make sentences understandable out of context, we introduce a QA-based decontextualisation model to enrich them with the necessary context. The experimental results show the superiority of our method over previous methods, including document-level claim extraction and evidence retrieval, as indicated by human evaluation and automatic evaluation. In future work, we plan to extend our document-level claim extraction method to extract salient check-worthy claims from multimodal web articles.

\section*{Acknowledgements}
\setlength{\intextsep}{0pt}
\setlength{\columnsep}{8pt}
\begin{wrapfigure}{l}{0.44\columnwidth}
    \includegraphics[width=0.44\columnwidth]{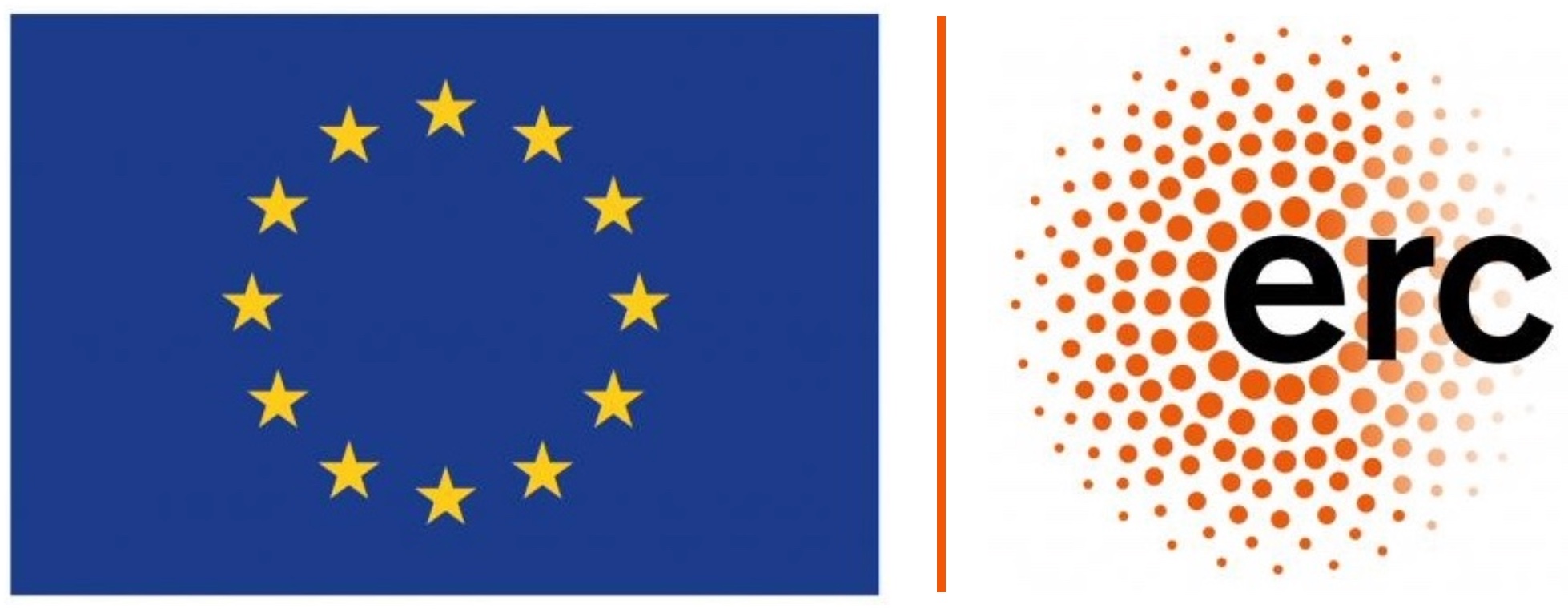} 
\end{wrapfigure}
This project has received funding from the European Research Council (ERC) under the European Union’s Horizon 2020 Research and Innovation programme grant AVeriTeC (Grant agreement No. 865958).  We thank David Corney from Full Fact for his help with evaluating the output of our models, and Yulong Chen for his helpful comments and discussions. We would also like to thank the anonymous reviewers for their helpful questions and comments that helped us improve the paper.

\section*{Limitations}
While our method has demonstrated superiority in extracting salient check-worthy claims and improving evidence retrieval, we recognize that our method is not able to decontextualise all ambiguous sentences, particularly those that lack the necessary context in the $source \ url$. Also, human fact-checkers have different missions, thus check-worthiness claims to one fact-checking organization may not be check-worthiness to another organization (\ie some organizations check parody claims or claims from satire websites, while others do not). 
% Thus retraining the claims from AVeriTeC may be required. 
Furthermore, since the documents we use are extracted from the $source \ url$, a powerful web scraper is required when pulling documents from $source \ urls$. Moreover, our method assumes salient claims are derived from central sentences. Although this assumption is true in most cases, it may be inconsistent with central claims collected by human fact-checkers. Besides, we use the chrF metric to calculate the similarity between claims extracted from the $source \ url$ and the gold claim, while gold claims are decontextualised by the fact-checkers with fact-checking articles and may contain information that is not in the original article, thus the metrics used to evaluate document-level claims are worth further exploring.

\section*{Ethics Statement}
We rely on fact-checks from real-world fact-checkers to develop and evaluate our models. Nevertheless, as any dataset, it is possible that it contains biases which influenced the development of our approach. Given the societal importance of fact-checking, we advise that any automated system is employed with human oversight to ensure that the fact-checkers fact-check appropriate claims.

\bibliography{custom}
\bibliographystyle{acl_natbib}

\appendix
\setcounter{table}{0}
\setcounter{figure}{0}
\setcounter{section}{0}
\setcounter{equation}{0}
\renewcommand{\thetable}{A\arabic{table}}
\renewcommand{\thefigure}{A\arabic{figure}}
\renewcommand{\thesection}{A\arabic{section}}
\renewcommand{\theequation}{A\arabic{equation}}

% \clearpage
% \onecolumn
\section{Statistic of Source URLs}
We described the statistic of the source URLs of the samples for document-level CE in Table \ref{appendix_A1}.
\\

\begin{table}[!h]
    \small
	\centering
	\begin{tabular} {l|c} \toprule
		\tabincell{c}{URL}  &\#$\rm sample$ \\  \midrule
    twitter.com     &241    \\     
    facebook.com    &235    \\
    perma.cc        &63    \\
    channelstv.com  &37    \\
    aljazeera.com   &34    \\
    president.go.ke &27    \\
    gov.za          &25    \\
    instagram.com   &18    \\
    c-span.org      &16    \\
    factba.se       &13    \\
    axios.com       &12    \\
    youtu.be        &12    \\
    rumble.com      &12    \\
    abcnews.go.com  &11    \\
    rev.com         &11    \\
    cnn.com         &10    \\
    news24.com      &10    \\
    punchng.com     &10    \\
    washingtonpost.com      &10    \\
    cbsnews.com     &8    \\
    foxnews.com     &8    \\
    misbar.com      &7    \\
    thegatewaypundit.com    &7    \\
    politifact.com  &7    \\
    nypost.com      &7    \\
    nbcnews.com     &7    \\
    telegraph.co.uk &7    \\
    wisn.com        &6    \\
    tatersgonnatate.com     &6    \\
    bustatroll.org  &6    \\
    dailymail.co.uk &5    \\
    whitehouse.gov  &5    \\
  \bottomrule
	\end{tabular}
	\caption{\footnotesize Statistic of the source URLs of the samples for document-level CE. We only list URLs with a total number number greater than 5. }
	\label{appendix_A1}
\end{table}

\begin{table*}[!h]
    % \small
	\centering
	\begin{tabular} {l| c | c | c | c | c | c} \toprule
         \multirow{2}{*}{Method} &\multicolumn{3}{c}{Sentence$^\star$} &\multicolumn{3}{c}{Dec. Sentence$^\star$}  \\ 
            \cmidrule(lr){2-4} \cmidrule(lr){5-7}
            &SARI &BERTScore &chrF &SARI &BERTScore &chrF \\  \midrule 
            Claimbuster   &6.23  &82.7  &24.3 &6.24  &82.8  &24.5 \\ \midrule
          Lead Sentence  &6.41 &83.4 &23.8 &-  &-  &- \\
            LSA           &5.56 &83.2  &24.1  &5.57 &83.2  &24.3 \\
            % KL-Sum        &6.00  &83.0  &23.5  &6.02 &83.0  &23.7 \\
            TextRank      &6.60  &83.1  &24.5  &6.61 &83.1  &25.4 \\ 
    %         Flan-T5      &-  &-  &19.3  &- &-  &-\\
		  % GPT-3.5-turbo &-  &-  &20.3  &- &-  &- \\ \midrule
            BertSum       &6.54  &83.6  &25.6  &6.55 &83.6  &25.9 \\ \midrule
            Ours         &6.56  &83.7  &25.9  &6.70 &83.8  &26.4 \\ 
  \bottomrule
	\end{tabular}
	\caption{\footnotesize  Results of Document-level CE on three different metrics.}
	% \vspace{-2mm}
	\label{appendix_A2}
\end{table*}

\section{Evaluation Metrics}
We use the following metrics to assess the similarity between the claim decontextualised by our method and the $claim$ decontextualised by fact-checkers.
$\bf SARI$ \cite{xu2016optimizing} is developed to compare the claim with the reference claim by measuring the goodness of words that are added, deleted and kept. $\bf BERTScore$ \cite{zhang2019bertscore} is utilized to compute the semantic overlap between the claim and the reference claim by sentence representation. 
Since most $claims$ in AVerTeC are decontextualised by fact-checkers with fact-checking articles, they may contain some information that is not in the $source \ url$, making it challenging for SARI and BERTScore to be used as evaluation metrics in this task. Thus, we use the chrF as our main evaluation metric for claim extraction.

\section{Implementation Details}
All models we use in this paper are pre-trained models (\eg BertSum) or approaches that do not require training (\eg BM25). The hyperparameters of each model can be found in the original paper. To help readers reproduce our method, we have released our code on GitHub\protect\footnotemark[5].

\footnotetext[5]{\url{https://github.com/Tswings/AVeriTeC-DCE}}

\section{ChatGPT for Decontextualisation}
To verify how well ChatGPT would do on decontextualisation, we use ChatGPT to decontextualise three claim sentences in Figure \ref{fig3}. The ChatGPT prompt for decontextualisation is as follows:

\begin{tcolorbox}[colback=white, colframe=black, title=ChatGPT Prompt]
Claim: \ \ \  [claim]      \\
Context: [context]  \\

To rewrite the Claim to be understandable out of context based on the Context, while retaining its original meaning.    \\
\rule{\linewidth}{1pt} 
Decontextualised sentences produced by ChatGPT:    \\
$\bullet$ Sentence 1: Barack Obama publicly expressed sympathy for ISIS and hinted at providing them with American resources during his 2008 State of the Union address.  \\
$\bullet$ Sentence 2: During a specific event, there was a delay in obtaining approval from a certain authority for assistance requested by the Capitol Police. \\
$\bullet$ Sentence 3: The Indian government lacks adequate storage facilities for managing the large surplus of grain it possesses. 
\end{tcolorbox}

% \noindent 
From the results, we can see that ChatGPT can produce well-formed claims that can be understood out of context, but it tends to rephrase the claim. In AVeriTeC (Appendix J.3.1), the decontextualised claims are required to be as close as possible to their original form. Our method tends not to change the original claims, but to rewrite only the ambiguous information units in claims, thus our generated claims are closer to the claims decontextualised by annotators than ChatGPT.

\end{document}